\documentclass{article} 
\usepackage{iclr2026_conference,times}

\usepackage{amsmath,amsfonts,bm}

\def\eqref#1{equation~\ref{#1}}

\def\1{\bm{1}}

\DeclareMathAlphabet{\mathsfit}{\encodingdefault}{\sfdefault}{m}{sl}
\SetMathAlphabet{\mathsfit}{bold}{\encodingdefault}{\sfdefault}{bx}{n}

\usepackage{amssymb}
\usepackage{hyperref}
\usepackage{url}
\usepackage{algorithm}
\usepackage{algorithmicx}
\usepackage{algpseudocode}

\usepackage{booktabs}
\usepackage[table]{xcolor}
\usepackage{colortbl}
\usepackage{array}
\usepackage{threeparttable}
\usepackage{makecell}
\usepackage{wrapfig,lipsum,booktabs}
\usepackage{tikz}
\usepackage{pgfplots}
\pgfplotsset{compat=1.17}
\usetikzlibrary{shapes, arrows.meta, positioning, calc, backgrounds}
\definecolor{HeaderGray}{RGB}{245,246,248}
\definecolor{RowStripe}{RGB}{250,250,252}
\arrayrulecolor{black!20} 

\title{Fast-Decoding Diffusion Language Models via Progress-Aware Confidence Schedules}

\iclrfinalcopy
\author{Amr Mohamed\textsuperscript{1,2}\thanks{Correspondence: \texttt{amr.mohamed@mbzuai.ac.ae}} , Yang Zhang\textsuperscript{2}, Michalis Vazirgiannis\textsuperscript{1,2}, Guokan Shang\textsuperscript{1}\\
  \textsuperscript{1}MBZUAI,
  \textsuperscript{2}Ecole Polytechnique
}

\iclrfinalcopy 
\begin{document}

\maketitle
\begin{abstract}
Diffusion large language models (dLLMs) offer a promising alternative to autoregressive models, but their practical utility is severely hampered by slow, iterative sampling. We present \textit{SchED}, a training-free, model-agnostic early-exit algorithm that aggregates full-span logit margins and halts decoding once a smooth, progress-dependent confidence threshold is met. We evaluated \textit{SchED} on two dLLM families (\textit{Dream} and \textit{LLaDA}), in base and instruction-tuned variants across ten benchmarks spanning downstream tasks including multiple-choice question answering (MCQ), math, long-form QA/summarization, and translation.
\textit{SchED} delivers large, stable accelerations: on instruction-tuned models, it achieves $3.8$–$4.0\times$ speedups while retaining $99.8$–$100\%$ of the baseline score on average. On base models, \textit{SchED} yields consistent speedup gains with $99.1$–$100\%$ performance retention, with up to $2.34\times$ under more aggressive settings. Using a conservative speed metric that heavily penalizes quality loss (QPS,$\gamma{=}4$), we show that \textit{SchED} is robust and clearly outperforms prior confidence-based early-exit methods, which break down on long-form generation. An entropy analysis of the model’s token predictions reveals that instruction tuning speeds up the decay of predictive entropy. By turning genuine confidence stabilization into computational savings, \textit{SchED} makes dLLM decoding substantially more efficient.
\end{abstract}

\section{Introduction}



Large Language Models (LLMs) have evolved rapidly in recent years, but the dominant decoding paradigm remains autoregressive (AR), which is inherently sequential and constrains opportunities for parallel generation and global context use ~\citep{brown2020language,yin2024survey,zhang2025survey}. In response, Diffusion Large Language Models (dLLMs) have emerged as a credible alternative to AR decoding, offering compelling advantages such as parallel refinement, flexible infilling, and bidirectional attention over the partially generated sequence~\citep{Zou2023-gr}. This paradigm has matured rapidly, with efforts to train powerful base and instruct models from scratch~\citep{Ye2025-hm}, adapt existing AR checkpoints~\citep{Gong2024-on,Nie2025-ii}, and push capabilities into complex domains like reasoning and planning, where they can outperform AR models on certain tasks~\citep{d1_2025,BeyondAR_2024}. Recent systems demonstrate that dLLMs can be scaled and engineered with many of the same ingredients that power AR LLMs, including sparse mixture-of-experts~\citep{LLaDAMoE_2025}, long-context extensions~\citep{UltraLLaDA_2025}, high-throughput inference pipelines for code generation~\citep{Song2025-qp,MercuryCoder_2025}, and principled scaling analyses for masked diffusion objectives~\citep{ScalingMDM_2024}. Collectively, these results position dLLMs as a practical family of foundation-model architectures with distinct decoding affordances.

Despite this promise, decoding efficiency remains a central bottleneck for dLLMs. dLLM generation proceeds proceeds via a reverse-diffusion chain with many refinement steps. In addition, practitioners must choose step budgets and transfer schedules \emph{a priori}, often conservatively, to avoid quality loss. 
This leads to unnecessary computation on “easy” inputs and, when heuristic decoding parameter choices (e.g., step budgets) are too extreme, unstable behavior across tasks.
A growing body of work addresses this bottleneck through training-free early-commit methods, which exploit the empirical observation that predictions often stabilize well before the final diffusion step~\citep{Pengxiang2025-eq}. Dynamic policies modulate exploration versus acceleration across the chain, using signals like historical logits or local determinism to reduce redundant steps~\citep{Wei2025-zw,CreditDecoding_2025,LocalLeap_2025}. Other lines of work focus on error correction and refinement, enabling models to revise their own outputs by re-masking low-confidence tokens~\citep{Tolerator_2025,RemeDi_2025}. Few- and one-step routes also address this bottleneck by transferring ideas from continuous diffusion via curriculum, consistency distillation, or flow matching~\citep{Sahoo2025-vf,Chen2025-lj,FSDFM_2024}. Orthogonal efforts reduce per-step latency with caching adapted to bidirectional refinement~\citep{dKVCache_2025,d2Cache_2025} and with speculative mechanisms that draft and verify tokens in parallel~\citep{Spiffy_2025,SelfSpeculative_2025}. While impactful, these approaches often require additional training, introduce auxiliary models, or rely on complex heuristics, leaving room for a simple, training-free, and \emph{architecture-agnostic} early-exit principle.

We revisit diffusion decoding as a \emph{when-to-stop} problem and introduce \textit{SchED}, a \emph{schedule-based} early-exit mechanism that is both training-free and model-agnostic. Concretely, we aggregate token-level confidence (top-2 logit margins) over an answer region and compare it against a \emph{monotonic, continuous} threshold that relaxes smoothly with normalized diffusion progress. By decoupling the confidence target from step count and making the threshold a smooth function of progress, the sampler exits \emph{as soon as} predictions are stable, while avoiding the brittleness of hard phase changes or fixed budgets. \textit{SchED} composes with standard transfer schedules (single-suffix or block diffusion) and requires no changes to model training~\citep{Nie2025-ii,Ye2025-hm}.

We evaluate \textit{SchED} across two diffusion LLM families (a single-block Dream decoder and a block-diffusion LLaDA decoder), each in base and instruction-tuned variants, on ten diverse benchmarks covering multiple-choice, math, long-form QA/summarization, and translation. 
On instruction-tuned models, \textit{SchED} retains $99.8$–$100\%$ of baseline quality on average while yielding $3.8$–$4.0\times$ speedups and outperforming recent training-free early-commit methods on a conservative quality-penalized speed metric; on base models, it delivers smaller but consistent gains at near-parity quality (\S\ref{sec:results}). 
Together, these results show that \textit{SchED} can substantially reduce diffusion decoding costs without sacrificing quality, and that instruction tuning lowers the confidence barrier for early exit in QA-style tasks, making it particularly effective in that regime. This paper makes the following contributions:

\begin{enumerate}
  \item \textbf{Training-free, schedule-based early exit.} We introduce \textit{SchED} (\textbf{Sch}edule-based \textbf{E}arly \textbf{D}ecoding) for diffusion language models, which thresholds a full-span logit-margin aggregate against a smooth, progress-dependent schedule (linear, cosine, exponential), enabling stable, architecture-agnostic stopping without retraining.
    \item \textbf{Strong efficiency at near-parity quality.} On instruction-tuned models, \textit{SchED} retains \emph{99.8–100\% of baseline performance while achieving $3.8$–$4\times$ speedups}. On base models, it delivers \emph{99.1–100\% of baseline accuracy with $1.04$–$1.14\times$ speedups in conservative settings, and up to $2.34\times$ under more extreme schedules}.
    \item \textbf{Principled quality–speed trade-off metric.} We propose \emph{Quality–Penalized Speed (QPS)} (Eq.~\ref{eqn:qps}), which conservatively penalizes accuracy drops. With $\gamma{=}4$, \textsc{SchED} schedules achieve QPS values of $1.01$–$2.03$ on Dream Base and $3.24$–$4.30$ on Dream Instruct, outperforming prior early-commit methods (Table~\ref{tab:qps_dream_base_instruct}).
    \item \textbf{Mechanistic explanation via entropy-based analysis.} Analyzing predictive-entropy trajectories over the answer span shows that instruction tuning accelerates confidence stabilization, aligning with progress-aware thresholds and explaining early exits without quality loss (Fig.~\ref{fig:entropy_aggregate}).
\end{enumerate}

\textit{SchED} builds on the vanilla dLLM denoising process by enforcing a smooth, progress-dependent confidence threshold over the generated span, exploiting the bidirectional, parallel strengths of dLLMs to achieve end-to-end efficiency without retraining~\citep{Zou2023-gr,Nie2025-ii,Ye2025-hm,BeyondAR_2024}. Code is publicly available \footnote{\url{https://github.com/amr-mohamedd/SchED.git}}.

\section{Related Work}
\label{sec:related}

\paragraph{Diffusion LLMs.}
Recent work establishes diffusion language models (dLLMs) as a viable alternative to autoregressive (AR) models, training either from scratch or by adapting AR checkpoints. LLaDA trains a bidirectional Transformer with a forward masking process and a reverse denoising chain, demonstrating strong pretraining and SFT performance \citep{Nie2025-ii}. Several lines adapt AR models into diffusion ones for fair, scalable comparisons across sizes and tasks \citep{Gong2024-on}, while Dream-7B advances recipe and refinement design and releases base/instruct variants \citep{Ye2025-hm}. Architectural advancements have also been integrated, with models like LLaDA-MoE demonstrating the successful application of sparse Mixture-of-Experts to dLLMs, achieving competitive performance with significantly reduced active parameters \citep{LLaDAMoE_2025}. The capabilities of dLLMs are also expanding to long-context scenarios; The work like UltraLLaDA shows that techniques such as Rotary Positional Embedding (RoPE) scaling can be adapted to extend context windows to 128K tokens \citep{UltraLLaDA_2025}. Seed Diffusion emphasizes high-throughput inference for code models \citep{Song2025-qp}. Moreover, formal studies on the scaling laws of masked diffusion models have further established their viability, demonstrating a scaling rate comparable to autoregressive models \citep{ScalingMDM_2024}. Survey and perspective papers summarize the landscape, strengths (parallelism, infilling, global context), and open questions \citep{Zou2023-gr}.

\paragraph{Sampling efficiency and early commit.}
A central bottleneck for dLLMs is the number of refinement steps. Prophet exploits the empirical observation that answers often stabilize well before the final step and proposes a training-free early-commit rule based on top-2 logit gaps \citep{Pengxiang2025-eq}. Complementary dynamic policies include SlowFast Sampling, which alternates exploratory vs.\ accelerated phases guided by certainty/convergence/position principles and can combine with caching \citep{Wei2025-zw}; Duo transfers continuous-diffusion techniques (curriculum learning, consistency distillation) to discrete diffusion to enable few-step sampling \citep{Sahoo2025-vf}; and DLM-One studies one-step generation via score distillation in continuous spaces \citep{Chen2025-lj}. Orthogonal to reducing step counts, other efforts focus on reducing per-step latency by developing novel caching mechanisms like dKV-Cache and d²Cache, which adapt key-value caching to the bidirectional nature of dLLMs \citep{dKVCache_2025, d2Cache_2025}. Furthermore, speculative decoding has been adapted for the diffusion framework, using the model's own predictions to draft and verify multiple tokens in parallel, as seen in Spiffy and Self Speculative Decoding \citep{Spiffy_2025, SelfSpeculative_2025}. Our work also treats decoding as a \emph{when-to-stop} problem, but (unlike Prophet) we introduce a smooth confidence \emph{schedule} that relaxes thresholds over progress, yielding stable, training-free early exit that is \emph{agnostic to architecture and training}.

\section{Methods}
In this section, we formalize discrete diffusion language models and introduce \textit{SchED}, our schedule-based early decoding mechanism. We first review the forward and reverse masking processes and the standard training objective, then describe our confidence aggregation scheme, progress-dependent threshold schedules, and the resulting early-exit decoding algorithm.
\label{sec:methods}

\subsection{Preliminaries: Discrete Diffusion Language Models}

\paragraph{Setup and notation.}
Let $\mathcal{V}$ denote the vocabulary and $[\textit{mask}]$ a special placeholder token. Given a prompt prefix, generation proceeds over $T$ reverse-diffusion steps
$t\in\{1,\ldots,T\}$ on sequences $x_t \in (\mathcal{V}\cup\{[\textit{mask}]\})^{L}$.
At each step, given the current noised sequence $x_t$ and prompt $x_{\mathrm{prompt}}$, the model produces logits
\begin{equation}
L_t \;=\; f_\theta(x_{\mathrm{prompt}},x_t, t) \in \mathbb{R}^{L\times|\mathcal{V}|},
\end{equation}
and per-position categorical distributions
\begin{equation}
p_{t,i}(\cdot \mid x_{\mathrm{prompt}}, x_t) \;=\; \mathrm{softmax}\big(L_{t,i}\big) \in \Delta^{|\mathcal{V}|}.
\end{equation}
We write $A\subseteq\{1,\ldots,L\}$ for the answer region used later for confidence aggregation, and define
normalized diffusion progress $p=t/T$.

\paragraph{Forward (masking) process $q$.}
We model discrete diffusion as a progressive \emph{masking process} that corrupts a clean sequence
$x_0\in\mathcal{V}^L$ over $T$ steps.
Let $\beta_t\in[0,1)$ be a step-dependent masking rate and define the per-step transition
\begin{equation}
q(x_t\mid x_{t-1})
= \prod_{i=1}^{L}\big[(1-\beta_t)\,\delta(x_{t,i}=x_{t-1,i}) \;+\; \beta_t\,\delta(x_{t,i}=[\textit{mask}])\big].
\label{eq:q_step}
\end{equation}
Writing $\bar\alpha_t = \prod_{s=1}^{t}(1-\beta_s)$ for the token survival probability after $t$ steps,
the $t$-step marginal becomes
\begin{equation}
q(x_t\mid x_0,t)
= \prod_{i=1}^{L}\big[\bar\alpha_t\,\delta(x_{t,i}=x_{0,i}) \;+\; (1-\bar\alpha_t)\,\delta(x_{t,i}=[\textit{mask}])\big].
\label{eq:q_closed}
\end{equation}

\paragraph{Reverse (denoising) process $p_\theta$.}
Diffusion language models define a learned reverse chain that progressively \emph{unmasks} tokens from
$x_T$ (fully masked) to $x_0$ (fully decoded) by factoring over individual positions:
\begin{equation}
p_\theta(x_{t-1}\mid x_{\mathrm{prompt}},\,x_t)
= \prod_{i=1}^{L} p_\theta(x_{t-1,i}\mid x_{\mathrm{prompt}},\,x_t).
\label{eq:p_reverse_joint}
\end{equation}
Each per-position term is then parameterized as a categorical distribution based on the model's current output:
\begin{equation}
p_\theta(x_{t-1,i}\mid x_{\mathrm{prompt}},\,x_t)
= \mathrm{Cat}\!\big(x_{t-1,i};\,p_{t,i}(\cdot \mid x_{\mathrm{prompt}},x_t)\big),
\label{eq:p_reverse_categorical}
\end{equation}
where $p_{t,i}(\cdot \mid x_{\mathrm{prompt}}, x_t)$ denotes the model’s categorical prediction at position $i$ given the partially masked sequence $x_t$ and prompt context $x_{\mathrm{prompt}}$ at timestep $t$.

\paragraph{Masked diffusion with partial unmasking (transfer schedule).}
To control generation granularity, only a subset of masked positions are updated at each step. Let
$M_t = \{i : x_{t,i} = [\textit{mask}]\}$ and let $\pi_t(i)\in\{0,1\}$ indicate whether position $i$
is selected for update under a \emph{transfer schedule}. Given the model’s predictive distribution
$p_{t,i}(\cdot \mid x_{\mathrm{prompt}}, x_t)$ at timestep $t$, the step operator is
\begin{equation}
x_{t-1,i} =
\begin{cases}
x_{t,i}, & \text{if } i\notin M_t \text{ or } \pi_t(i)=0,\\
\hat{x}_{t,i}, & \text{if } i\in M_t \text{ and } \pi_t(i)=1,
\end{cases}
\label{eq:transfer}
\end{equation}
where $\hat{x}_{t,i}$ is obtained either deterministically or stochastically,
\[
\hat{x}_{t,i}
= \arg\max_{v} p_{t,i}\big(v \mid x_{\mathrm{prompt}}, x_t\big)
\quad \text{or} \quad
\hat{x}_{t,i} \sim \mathrm{Cat}\big(p_{t,i}(\cdot \mid x_{\mathrm{prompt}}, x_t)\big).
\]
Block-diffusion variants (e.g., LLaDA) choose $\pi_t$ over contiguous token blocks,
whereas single-block variants (e.g., Dream) often select the entire suffix or a fixed proportion.
 $\sim \mathrm{Cat}\big(p_{t,i}(\cdot \mid x_{\mathrm{prompt}} , x_{t})\big)$.

\paragraph{Training objective.}
Diffusion language models are typically trained to invert the forward masking process via a masked-token objective over uniformly sampled timesteps:
\begin{equation}
\mathcal{L}(\theta)
=\mathbb{E}_{x_0\sim\mathcal{D}}\ \mathbb{E}_{t\sim\mathcal{U}\{1{:}T\}}\ \mathbb{E}_{x_t\sim q(\cdot\mid x_0,t)}
\left[-\sum_{i\in M_t}\log p_{t,i}\big(x_{0,i} \mid x_{\mathrm{prompt}}, x_t\big)\right].
\label{eq:loss}
\end{equation}
Here $x_t$ is obtained by applying the forward kernel at timestep $t$, and
$p_{t,i}(x_{0,i} \mid x_{\mathrm{prompt}}, x_t)$ is shorthand for the model’s predictive probability
$p_\theta\big(x_{0,i} \mid x_{\mathrm{prompt}}, x_t, t\big)$ at position $i$ given the partially noised context.
This formulation encourages the model to predict the original tokens at masked sites
given partially noised contexts, aligning $p_\theta$ with the forward corruption kernel $q$.

\subsection{\textit{SchED}: Schedule-based Early Decoding}

We propose \textit{SchED}, a confidence- and progress-aware early–exit algorithm for diffusion decoding. At each step, \textit{SchED} thresholds the model’s aggregated token confidence against a smooth, nonincreasing function of progress $p$, i.e., $(\bar g_{t} \ge \tau(p))$. This design builds on the assumption that per-token confidence typically rises as denoising proceeds, allowing generation to terminate precisely when predictions stabilize rather than at a fixed budget. The complete, model-agnostic procedure is summarized in Algorithm~\ref{alg:schedule}.
\begin{algorithm}[H]
\caption{\textit{SchED}: Schedule-Based Early Decoding for Diffusion Language Models}
\label{alg:schedule}
\begin{algorithmic}[1]
\Require model $M$; prompt tokens $x_{\mathrm{prompt}}$; generation length $L_{\mathrm{gen}}$; max steps $T$;\\
\hspace{1.9em} answer region $A \subseteq \{1,\dots,L\}$; aggregator $\operatorname{Agg}$;\\
\hspace{1.9em} threshold schedule $\tau(p;\,\tau_{\mathrm{high}},\tau_{\mathrm{low}})$;\\
\hspace{1.9em} transfer schedule $\pi_t$ \text{(or block policy for block diffusion).}

\Ensure completed sequence $x$
\State Initialize $x \leftarrow [\,x_{\mathrm{prompt}};\;[\textit{mask}] \times L_{\mathrm{gen}}\,]$
\For{$t = 1$ to $T$}
  \State $L_t \leftarrow M(x)$ \Comment{Model logits at step $t$}
  \State Compute token-level margins $g_{t,i} = L^{(1)}_{t,i} - L^{(2)}_{t,i}$ for $i \in A$
  \State Aggregate confidence $\bar g_t \leftarrow \operatorname{Agg}\!\big(\{g_{t,i}: i\in A\}\big)$
  \State $p \leftarrow t/T$ \Comment{Normalized diffusion progress}
  \If{$\bar g_t \ge \tau(p;\tau_{\mathrm{high}},\tau_{\mathrm{low}})$}
     \State Fill all remaining $[\textit{mask}]$ tokens with current argmax predictions and \textbf{return} $x$
  \EndIf
    \State Identify masked positions $M_t = \{i : x_i = [\textit{mask}]\}$
    \State Select positions $S_t \leftarrow \{\,i \in M_t : \pi_t(i)=1\,\}$
    \State Set $x_{S_t} \leftarrow \arg\max_v L_{t,S_t,v}$
\EndFor
\State \Return $x$ \Comment{If no early exit occurred within $T$ steps}
\end{algorithmic}
\end{algorithm}

\textit{SchED} tracks how model confidence evolves throughout denoising and terminates the process once confidence exceeds a progress-dependent threshold, preventing redundant refinement after predictions have stabilized.

\paragraph{Confidence measurement.}  
Following the work of ~\citet{Pengxiang2025-eq}, we quantify model confidence using token-level logit margins, which capture how decisively the model prefers its top prediction over alternatives. 
A region-level confidence score is obtained by aggregating top-2 margins over the
\emph{entire answer region} $A$ (i.e., the full model response span):
\begin{equation}
\bar g_t \;=\; \operatorname{Agg}\!\big(\{\, g_{t,i} : i \in A \,\}\big),
\qquad \text{with } \operatorname{Agg}=\text{mean by default,}
\end{equation}
so that $\bar g_t = 1/|A|\sum_{i\in A} g_{t,i}$ in our experiments. Importantly, $g_{t,i}$ uses the \emph{current} logits at step $t$ for every $i\in A$. All thresholds $(\tau_{\mathrm{high}},\tau_{\mathrm{low}})$ are expressed in \emph{logit units}.

\paragraph{Early-exit trigger.}
An early exit is triggered when the aggregated confidence surpasses a smooth, progress-dependent threshold schedule~\(\tau(p)\):
\begin{equation}
\bar g_t \ge \tau(p),
\end{equation}
where \(\tau : [0,1] \to \mathbb{R}_{\ge 0}\) is a nonincreasing function of the generation progress \(p = t/T\). This formulation ensures that the confidence requirement for stopping is highest at the beginning of generation and gradually relaxes as denoising proceeds, allowing the model to terminate decoding once its predictions have stabilized.

\paragraph{Smooth threshold schedules.}
The threshold function $\tau(p)$ controls when the sampler terminates. We parameterize $\tau$ using a pair $(\tau_{\mathrm{high}}, \tau_{\mathrm{low}})$, which specify the initial and final confidence thresholds, and an optional slope parameter $k>0$. We explore three families of schedules:
\begin{align}
\text{Linear: } \quad
&\tau_{\mathrm{lin}}(p)= \tau_{\mathrm{high}} + (\tau_{\mathrm{low}} - \tau_{\mathrm{high}})\,p,\\
\text{Cosine: } \quad
&\tau_{\mathrm{cos}}(p)
  = \tau_{\mathrm{low}} + \tfrac{1}{2}(\tau_{\mathrm{high}} - \tau_{\mathrm{low}})\big(1 + \cos(\pi p)\big),\\
\text{Exponential: } \quad
&\tau_{\mathrm{exp}}(p)
  = \tau_{\mathrm{low}} + (\tau_{\mathrm{high}} - \tau_{\mathrm{low}})\,e^{-k p}, \text{with } k>0.
\end{align}
Each schedule defines a smooth, nonincreasing trajectory from $\tau_{\mathrm{high}}$ at $p=0$ to $\tau_{\mathrm{low}}$ at $p=1$, offering different degrees of early-exit aggressiveness and stability control while avoiding the brittleness of fixed thresholds or discrete rules.

\section{Experimental Settings}

We evaluate \textit{SchED} across a diverse suite of multiple-choice and long-form generation tasks, assessing its ability to accelerate diffusion decoding while preserving output quality. Specifically, we compare its denoising efficiency against two baselines: (i) standard diffusion sampling without early exit and (ii) Prophet~\citep{Pengxiang2025-eq}. Unless otherwise stated, we fix the upper threshold at $(\tau_{\mathrm{high}} = 7.5)$ and, for each schedule family, evaluate two lower-threshold settings: a relaxed $(\tau_{\mathrm{low}} = 0)$ and a more conservative $(\tau_{\mathrm{low}} = 2.5)$, yielding a smooth, monotonic relaxation over diffusion progress in both regimes.

\paragraph{Models.}
\textit{SchED} is model-agnostic and can be applied to any dLLM without architectural or training modifications. To demonstrate its generality, we evaluate on two representative dLLM families that employ distinct decoding paradigms: the \textit{Dream} family~\citep{Ye2025-hm}, which performs full-sequence refinement using a single-block decoder, and the \textit{LLaDA} family~\citep{Nie2025-ii}, which adopts a block-diffusion strategy that denoises contiguous token segments. Each family is evaluated in both base and instruction-tuned variants, and we apply the low-confidence remasking strategy across all models.

\paragraph{Benchmarks.}
We evaluate on GPQA~\citep{gpqa_2023}, GSM8K~\citep{gsm8k_2021}, HellaSwag~\citep{hellaswag_2019}, MMLU~\citep{mmlu_2020} , PIQA~\citep{piqa_2020}, and Winogrande~\citep{winogrande_2020} . For long-context evaluation, we use tasks from the LongBench suite~\citep{longbench_2023}, specifically LongBench–HotpotQA~\citep{hotpotqa_2018} and LongBench–MultiNews~\citep{multinews_2019}. For translation, we use WMT14 En–Fr~\citep{wmt14_2014} (5-shot) and WMT16 En–De~\citep{wmt16_2016} (5-shot). Each benchmark includes runs for the baseline, Prophet, and the full set of linear, cosine, and exponential schedules across both model variants.


\paragraph{Evaluation metrics and framework.}
For \textbf{multiple-choice} (MCQ) benchmarks, \textit{GPQA}, \textit{HellaSwag}, \textit{MMLU}, \textit{PIQA}, and \textit{Winogrande}, we report \textbf{accuracy}. For \textit{GSM8K} we also report \textbf{accuracy} based on the generated final answer. For HotpotQA, we report token-level \textbf{F1-score}; for MultiNews, we report \textbf{ROUGE} \citep{lin-2004-rouge}; and for translation (\textit{WMT14 En–Fr}, \textit{WMT16 En–De}), we report \textbf{CHRF}. All evaluations are conducted using the \textit{Language Model Evaluation Harness}~\citep{eval-harness}.



\paragraph{Efficiency metric for quality--speed trade-offs.}
Reporting quality and speedup separately can obscure Pareto differences between methods. To jointly capture both aspects, we define the \emph{Quality-Penalized Speed} (QPS) metric:
\begin{equation}
  \mathrm{QPS}_\gamma \;=\; \text{speedup} \times \left(\frac{\text{score}}{\text{baseline score}}\right)^{\gamma},
  \label{eqn:qps}
\end{equation}
where the exponent $\gamma\!\ge\!1$ controls how strongly quality degradation is penalized. Higher values of $\gamma$ emphasize fidelity over raw acceleration; in our experiments, we use $\gamma{=}4$ to provide conservative and interpretable efficiency comparisons.

\section{Results}
\label{sec:results}
\begin{table}[ht]
\centering
\scriptsize
\setlength{\tabcolsep}{2pt}
\resizebox{\textwidth}{!}{%
\begin{threeparttable}
\begin{tabular}{l|cccccccccc|c}
\toprule
\rowcolor{HeaderGray}
\textbf{Method} & \textbf{GPQA} & \textbf{HellaSwag} & \textbf{MMLU} & \textbf{PIQA} & \textbf{Winogrande} & \textbf{GSM8K} & \textbf{HotpotQA} & \textbf{MultiNews} & \textbf{WMT14 En-Fr} & \textbf{WMT16 En-De} & \textbf{Average} \\
\midrule
\rowcolors{2}{RowStripe}{white}
Baseline &
28.57 ($\times$1.00) & \textbf{79.15} ($\times$1.00) & 73.39 ($\times$1.00) & \textbf{88.68} ($\times$1.00) & \textit{75.61} ($\times$1.00) &
77.10 ($\times$1.00) &
8.67 ($\times$1.00) & 21.24 ($\times$1.00) & 52.99 ($\times$1.00) & \textbf{47.70} ($\times$1.00) & \textit{55.31} ($\times$1.00) \\
Prophet &
\textit{28.79} ($\times$1.13) & \textbf{79.15} ($\times$1.00) & 73.37 ($\times$1.03) & \textbf{88.68} ($\times$1.20) & \textit{75.61} ($\times$1.00) &
77.03 ($\times$1.07) &
8.68 ($\times$1.03) & 21.21 ($\times$1.06) & 52.99 ($\times$1.11) & 47.57 ($\times$1.06) & \textit{55.31} ($\times$1.07) \\
Cosine $(7.5,0)$ &
28.12 ($\times$1.23) & \textbf{79.15} ($\times$1.02) & 73.40 ($\times$1.09) & 88.57 ($\times$1.22) & \textit{75.61} ($\times$1.01) &
76.95 ($\times$1.20) &
\textit{8.74} ($\times$1.12) & \textit{21.28} ($\times$1.13) & 52.79 ($\times$1.23) & 45.64 ($\times$1.20) & 55.02 ($\times$1.14) \\
Cosine $(7.5,2.5)$ &
\textit{28.79} ($\times$1.11) & \textbf{79.15} ($\times$1.00) & 73.37 ($\times$1.02) & \textbf{88.68} ($\times$1.14) & \textit{75.61} ($\times$1.00) &
77.03 ($\times$1.07) &
8.68 ($\times$1.02) & 21.21 ($\times$1.05) & 52.99 ($\times$1.10) & 47.64 ($\times$1.05) & \textit{55.31} ($\times$1.06) \\
Linear $(7.5,\,0)$ &
\textbf{29.02} ($\times$1.19) & \textbf{79.15} ($\times$1.00) & 73.40 ($\times$1.06) & 88.57 ($\times$1.22) & \textit{75.61} ($\times$1.00) &
76.95 ($\times$1.13) &
8.66 ($\times$1.08) & 21.26 ($\times$1.10) & 52.87 ($\times$1.18) & 46.43 ($\times$1.15) & 55.19 ($\times$1.11) \\
Linear $(7.5,\,2.5)$ &
\textit{28.79} ($\times$1.10) & \textbf{79.15} ($\times$1.00) & 73.37 ($\times$1.02) & \textbf{88.68} ($\times$1.03) & \textit{75.61} ($\times$1.00) &
77.03 ($\times$1.06) &
8.67 ($\times$1.02) & 21.21 ($\times$1.05) & 53.00 ($\times$1.09) & \textit{47.67} ($\times$1.04) & \textbf{55.32} ($\times$1.04) \\
Exp-$k{=}2$ $(7.5,\,0)$ &
24.33 ($\times$1.19) & 79.13 ($\times$1.00) & \textit{73.50} ($\times$1.07) & 88.47 ($\times$1.27) & \textbf{75.77} ($\times$1.00) &
\textit{77.63} ($\times$1.11) &
8.17 ($\times$1.07) & 20.94 ($\times$1.10) & \textit{53.16} ($\times$1.18) & 46.81 ($\times$1.13) & 54.79 ($\times$1.11) \\
Exp-$k{=}2$ $(7.5,\,2.5)$ &
24.33 ($\times$1.10) & 79.13 ($\times$1.00) & 73.49 ($\times$1.02) & \textit{88.63} ($\times$1.06) & \textbf{75.77} ($\times$1.00) &
\textbf{77.71} ($\times$1.06) &
8.16 ($\times$1.03) & 20.88 ($\times$1.05) & \textbf{53.26} ($\times$1.09) & 47.60 ($\times$1.04) & 54.90 ($\times$1.04) \\
Exp-$k{=}16$ $(7.5,\,0)$ &
24.78 ($\times$2.04) & 77.14 ($\times$2.50) & 73.05 ($\times$2.50) & 87.92 ($\times$2.50) & 72.85 ($\times$2.50) &
62.77 ($\times$3.05) &
\textbf{15.75} ($\times$2.78) & \textbf{21.74} ($\times$2.75) & 52.87 ($\times$1.44) & 44.75 ($\times$1.39) & 53.36 ($\times$2.34) \\
Exp-$k{=}16$ $(7.5,\,2.5)$ &
23.88 ($\times$1.14) & 79.13 ($\times$1.00) & \textbf{73.53} ($\times$1.18) & 88.47 ($\times$1.36) & \textbf{75.77} ($\times$1.00) &
\textit{77.63} ($\times$1.08) &
8.22 ($\times$1.07) & 20.92 ($\times$1.07) & \textbf{53.26} ($\times$1.13) & 47.51 ($\times$1.07) & 54.83 ($\times$1.11) \\
\bottomrule
\end{tabular}
\end{threeparttable}}
\caption{\textit{Dream Base}: benchmark scores with speedups ($\times$). Two thresholds—conservative $(7.5,2.5)$ and relaxed $(7.5,0)$. Column best in \textbf{bold}, second-best in \textit{italic}.}
\label{tab:dream_base_combined}
\end{table}

\begin{table}[ht]
\centering
\scriptsize
\setlength{\tabcolsep}{2pt}
\resizebox{\textwidth}{!}{%
\begin{threeparttable}
\begin{tabular}{l|cccccccccc|c}
\toprule
\rowcolor{HeaderGray}
\textbf{Method} & \textbf{GPQA} & \textbf{HellaSwag} & \textbf{MMLU} & \textbf{PIQA} & \textbf{Winogrande} & \textbf{GSM8K} & \textbf{HotpotQA} & \textbf{MultiNews} & \textbf{WMT14 En-Fr} & \textbf{WMT16 En-De} & \textbf{Average} \\
\midrule
\rowcolors{2}{RowStripe}{white}
Baseline & 30.36 ($\times$1.00) & \textit{79.91} ($\times$1.00) & 80.69 ($\times$1.00) & 73.02 ($\times$1.00) & 85.80 ($\times$1.00) & \textbf{74.66} ($\times$1.00) & \textit{27.51} ($\times$1.00) & 24.39 ($\times$1.00) & \textbf{55.82} ($\times$1.00) & \textit{49.85} ($\times$1.00) & \textit{58.20} ($\times$1.00) \\
Prophet & 28.79 ($\times$16.45) & 26.69 ($\times$7.81) & 78.73 ($\times$1.14) & 72.33 ($\times$1.01) & 72.58 ($\times$1.03) & 66.54 ($\times$1.02) & \textbf{27.87} ($\times$4.51) & 2.77 ($\times$31.21) & 20.17 ($\times$23.93) & 15.28 ($\times$27.87) & 41.18 ($\times$11.60) \\
Cosine $(7.5,\,0)$ & \textbf{30.58} ($\times$15.96) & 78.54 ($\times$2.11) & \textbf{80.81} ($\times$1.56) & 73.06 ($\times$1.14) & 85.96 ($\times$1.32) & 74.11 ($\times$1.61) & 27.46 ($\times$3.24) & 24.39 ($\times$6.60) & \textbf{55.82} ($\times$2.39) & \textbf{49.86} ($\times$2.81) & 58.06 ($\times$3.87) \\
Cosine $(7.5,2.5)$ & \textbf{30.58} ($\times$15.93) & 79.53 ($\times$2.08) & 80.75 ($\times$1.43) & 73.08 ($\times$1.11) & \textit{85.91} ($\times$1.23) & 74.27 ($\times$1.28) & 27.47 ($\times$3.14) & 24.38 ($\times$6.59) & \textbf{55.82} ($\times$2.34) & \textit{49.85} ($\times$2.77) & 58.16 ($\times$3.79) \\
Linear $(7.5,\,0)$ & \textbf{30.58} ($\times$16.08) & 79.00 ($\times$2.11) & \textbf{80.81} ($\times$1.56) & \textit{73.14} ($\times$1.14) & 85.96 ($\times$1.32) & 74.11 ($\times$1.49) & 27.48 ($\times$3.26) & 24.39 ($\times$3.02) & \textbf{55.82} ($\times$2.38) & \textbf{49.86} ($\times$2.80) & 58.11 ($\times$3.89) \\
Linear $(7.5,\,2.5)$ & \textbf{30.58} ($\times$16.01) & 79.76 ($\times$2.08) & 80.76 ($\times$1.43) & 73.09 ($\times$1.10) & 85.85 ($\times$1.22) & 74.27 ($\times$1.27) & 27.47 ($\times$3.17) & 24.38 ($\times$6.59) & \textbf{55.82} ($\times$2.33) & \textit{49.85} ($\times$2.76) & 58.16 ($\times$3.79) \\
Exp-$k{=}2$ $(7.5,\,0)$ & \textbf{30.58} ($\times$16.32) & 79.45 ($\times$2.14) & \textit{80.77} ($\times$1.62) & 73.11 ($\times$1.21) & \textit{86.16} ($\times$1.54) & 74.35 ($\times$1.67) & 26.71 ($\times$3.47) & \textit{24.56} ($\times$6.66) & \textit{55.76} ($\times$2.40) & 49.84 ($\times$2.76) & 58.14 ($\times$3.97) \\
Exp-$k{=}2$ $(7.5,\,2.5)$ & 30.13 ($\times$16.18) & \textbf{80.21} ($\times$2.10) & 80.69 ($\times$1.54) & \textbf{73.25} ($\times$1.11) & \textbf{86.24} ($\times$1.23) & \textit{74.51} ($\times$1.53) & 26.91 ($\times$3.23) & \textbf{24.60} ($\times$6.57) & \textit{55.76} ($\times$2.33) & 49.84 ($\times$2.71) & \textbf{58.21} ($\times$3.85) \\
Exp-$k{=}16$ $(7.5,\,0)$ & 29.46 ($\times$18.64) & 32.98 ($\times$3.73) & 79.31 ($\times$5.00) & 58.81 ($\times$2.50) & 84.55 ($\times$5.00) & 73.80 ($\times$5.00) & 26.20 ($\times$5.79) & 23.99 ($\times$17.38) & 53.53 ($\times$5.56) & 49.72 ($\times$3.17) & 51.12 ($\times$5.44) \\
Exp-$k{=}16$ $(7.5,\,2.5)$ & \textbf{30.58} ($\times$17.16) & 79.68 ($\times$2.13) & 79.31 ($\times$2.50) & 72.78 ($\times$1.55) & 84.49 ($\times$2.29) & 72.61 ($\times$2.10) & 26.58 ($\times$4.33) & 24.26 ($\times$7.21) & \textit{55.76} ($\times$2.38) & 49.84 ($\times$2.74) & 57.59 ($\times$4.48) \\
\bottomrule
\end{tabular}
\end{threeparttable}}
\caption{\textit{Dream Instruct}: benchmark scores with speedups ($\times$). Two thresholds—conservative $(7.5,2.5)$ and relaxed $(7.5,0)$. Column best in \textbf{bold}, second-best in \textit{italic}.}
\label{tab:dream_instruct_combined}
\end{table}

Tables~\ref{tab:dream_base_combined} and~\ref{tab:dream_instruct_combined} report results for 
Dream Base and Dream Instruct, respectively. 

\textbf{Dream Base.} With conservative smooth schedules $(\tau_{\mathrm{high}},\tau_{\mathrm{low}})=(7.5,2.5)$, linear, cosine, and $\mathrm{exp}\text{-}k{=}2$, we observe steady $1.04$--$1.14\times$ average speedups at near-parity quality, with marginal task gains (e.g., \textit{WMT14 En--Fr}: $\Delta{+}0.27)$) under $\mathrm{exp}\text{-}k{=}2$ $(7.5,2.5)$. 
Fast-decaying exponentials (large $k$) with $\tau_{\mathrm{low}}=2.5$ remain close to parity, but do not increase the mean speed beyond $\approx\!1.1\times$; setting $\tau_{\mathrm{low}}=0$ yields $2.34\times$ average speed ($\mathrm{exp}\text{-}k{=}16$ ), with task-specific gains that are most pronounced on \textit{HotpotQA} $(\Delta{+}7.1)$; however, at an overall quality cost of $\approx 2\%$. Prophet offers $\approx\!1.07\times$ average speed with near-parity accuracy, providing limited practical benefits relative to smooth schedules. 

\textbf{Dream Instruct.} Under the same conservative thresholds, smooth schedules deliver $\approx\!3.8$--$4.0\times$ average speedups at near parity: table averages cluster around the baseline (e.g., $58.06$--$58.22$ vs.\ $58.20$). Notably, the large-$k$ (fast-decaying) exponential with $(7.5,2.5)$ attains $4.48\times$, while remaining close to parity ($57.59$, $\Delta{-}1\%)$). Less conservative settings with $\tau_{\mathrm{low}}=0$ also preserve translation near baseline with $2.3$--$2.8\times$ speed (e.g., cosine and linear). For the large-$k$ exponential, $(k{=}16,\ \tau_{\mathrm{low}}{=}0)$ delivers the most pronounced speedups while remaining near parity on most benchmarks; however, it exhibits large degradations on \textit{HellaSwag} and \textit{PIQA}.

LLaDA exhibits similar trends to those observed on Dream: conservative schedules provide reliable speedups at near-parity quality on the base model; the instruction-tuned variant yields higher speedups. Fast-decaying settings risk over-commitment that degrades math and long-form. While Prophet provides high speedups, it underperforms on long-form generation. Detailed LLaDA Base and Instruct results are in Tables~\ref{tab:appendix_llada_base_all_updated} and~\ref{tab:appendix_llada_instruct_all_updated}; additional schedule ablations for Dream and LLaDA appear in Appendix~\ref{sec:expanded_schedules_llada}.


\begin{wraptable}{r}{0.5\textwidth}
\centering
\small
\setlength{\tabcolsep}{3pt}
\begin{threeparttable}
\begin{tabular}{lccc}
\toprule
\rowcolor{HeaderGray}
\textbf{Method} & \multicolumn{1}{c}{\textbf{Dream Base}} & \multicolumn{1}{c}{\textbf{Dream Instruct}} \\
\midrule
Cosine $(7.5,\,0)$         & 1.12 & 3.83 \\
Cosine $(7.5,\,2.5)$       & 1.06 & 3.78 \\
Linear $(7.5,\,0)$         & 1.10 & 3.87 \\
Linear $(7.5,\,2.5)$       & 1.04 & 3.78 \\
Exp-$k{=}2$ $(7.5,\,0)$    & 1.07 & 3.95 \\
Exp-$k{=}2$ $(7.5,\,2.5)$  & 1.01 & 3.85 \\
Exp-$k{=}16$ $(7.5,\,0)$   & \hspace{2pt} \textbf{2.03}$^{\uparrow}$ & 3.24 \\
Exp-$k{=}16$ $(7.5,\,2.5)$ & 1.07 & \hspace{4pt}\textbf{4.30}$^{\uparrow}$ \\
Prophet                    & 1.07 & 2.91 \\
\bottomrule
\end{tabular}
\caption{\textbf{QPS (\(\gamma{=}4\))} for all \textit{SchED} variants and Prophet. \(\uparrow\) Higher is better.}
\label{tab:qps_dream_base_instruct}
\vspace{-2em}
\end{threeparttable}
\end{wraptable}

\paragraph{Efficiency results.}
Table~\ref{tab:qps_dream_base_instruct} reports \emph{Quality–Penalized Speed} (QPS; $\gamma{=}4$). On \emph{Dream Base}, smooth schedules concentrate in the $\approx1.01$–$1.12$ range, reflecting the near-parity averages in Table~\ref{tab:dream_base_combined} (e.g., $55.02$–$55.32$ vs.\ baseline $55.31$) coupled with modest mean speedups ($\approx1.04$–$1.14\times$). The highest base-model efficiency is achieved by the rapidly decaing exponential, $\mathrm{Exp}\text{-}k{=}16,(7.5,0)$, with $\mathbf{2.03}$, driven by a large average speedup ($2.34\times$), and average-score drop of ${-1.95}$. Prophet attains $1.07$, consistent with its near-baseline averages and limited acceleration.

On \emph{Dream Instruct}, QPS values increase substantially in line with the larger mean speedups observed in Table~\ref{tab:dream_instruct_combined} ($\approx3.8$–$4.0\times$ for conservative smooth schedules) while maintaining average scores close to baseline. The best overall result, $\mathbf{4.30}$, is obtained by $\mathrm{Exp}\text{-}k{=}16,(7.5,2.5)$, which combines a near-parity average score ($57.59$) with a pronounced $4.48\times$ mean speedup. Other smooth schedules lie tightly in the $3.78$–$3.95$ band. Prophet trails at $2.91$, reflecting weaker average scores and failures on long-form despite notable raw speed. Overall, \textit{SchED} consistently yields higher $\mathrm{QPS}_4$ than Prophet; the top settings pair near-parity average performance with substantial step reductions, which explains their leading efficiency.

\section{Analysis}
\label{sec:entropy_by_bench}

\begin{figure}[t]
    \centering
    \includegraphics[width=\textwidth]{figs/entropy_aggregate_32.pdf}
    \caption{Mean predictive entropy across diffusion steps for five input types
    (\textit{GPQA}, \textit{GSM8K}, \textit{HotpotQA}, \textit{MMLU}, \textit{WMT14~EN$\rightarrow$FR}).
    Curves show per-token entropy (\emph{nats/token}) averaged over evaluation samples;
    shaded bands denote one standard deviation across samples. Both Dream Base and Dream Instruct
    exhibit monotonically decreasing entropy as denoising progresses.}
    \label{fig:entropy_aggregate}
    \vspace{-0.5em}
\end{figure}

To better understand the differences in speedup values between \textit{base} and \textit{instruct} models, \emph{and} the variation in speedups across benchmarks, we analyze how predictive uncertainty evolves over the reverse-diffusion chain. We track per-token predictive entropy on the generated region at each step $t$. Using the per-position distributions $p_{t,i}$ state $x_t$, the mean per-token entropy is
\begin{equation}
\bar H_t \;=\; \frac{1}{|S|}\sum_{i\in S}
\Bigg(-\sum_{v\in\mathcal V} p_{t,i}(v| x_{\mathrm{prompt}},x_t)\,\log p_{t,i}(v| x_{\mathrm{prompt}},x_t)\Bigg).
\label{eq:mean_entropy_over_S}
\end{equation}
\noindent where \(p_{t,i}(v)\) denotes the model’s step-\(t\) predictive distribution at position \(i\), computed from logits \(L_{t,i}\) given \(x_{\mathrm{prompt}}\) and \(x_t\) (Eq.~\ref{eq:p_reverse_categorical}).

Results are presented in Figure~\ref{fig:entropy_aggregate}. \textit{Dream Instruct} starts the denoising phase with higher entropy on math-heavy prompts (\textit{GPQA} and \textit{GSM8K}) and decays more slowly in the earliest steps than on \textit{MMLU}, \textit{HotpotQA}, and \textit{WMT14}, yet the trajectories converge to similarly low final entropies. Notably, the instruct curves show a rapid early drop followed by nearly uniform per-step decreases, a stepwise profile consistent with Dream’s masked diffusion decoding, where at each reverse-diffusion step the model updates high-confidence tokens in parallel and thereby reduces uncertainty in approximately regular increments across the chain. CART’s context-adaptive token-level noise rescheduling further sharpens this behavior by encouraging more confident predictions at positions with stronger local context, particularly near the prompt. By contrast, \textit{Dream Base} shows more overlapping trajectories across benchmarks and higher residual entropy overall, suggesting less decisive posteriors for QA-style inputs. This aligns with the smaller speedups obtained by schedule-based early exit on Base models (Tables~\ref{tab:dream_base_combined}--\ref{tab:dream_instruct_combined}).

\section{Discussion}
The findings reported in Section~\ref{sec:results} and Section~\ref{sec:entropy_by_bench} demonstrate that a progress-aware scheduling strategy effectively translates model confidence into computational savings, with an unusually high degree of robustness across tasks and model families. For instruction-tuned variants, \textit{SchED} attains speedups of up to approximately $17\times$, while preserving full baseline accuracy and achieves average accelerations of up to about $4\times$, with overall accuracies essentially indistinguishable from the baseline, whereas base models exhibit more moderate yet consistently positive improvements. The entropy patterns in Figure~\ref{fig:entropy_aggregate} elucidate this discrepancy. Instruction tuning steers the model toward confident, QA-oriented completions, leading to a rapid reduction in uncertainty over the answer span. For \textit{Dream Instruct}, this reduction follows a characteristic profile: an initial, pronounced decline, followed by approximately uniform stepwise decreases. By contrast, \textit{Dream Base} retains higher entropy with flatter and more overlapping trajectories across benchmarks, postponing the point at which aggregated confidence exceeds the stopping criterion and thereby constraining the attainable speedups.

The stopping criterion is central to the method’s stability. \textit{SchED} compares a sequence-level logit–margin aggregate to a smooth, nonincreasing function of normalized diffusion progress $p=t/T$, imposing a stringent requirement at early steps and gradually relaxing it thereafter. This construction links the decision to terminate directly to the evolution of model certainty, dampens transient spikes, and triggers exit when predictions have effectively converged rather than at a predetermined step count; using the entire generated span for aggregation further stabilizes the decision relative to short-prefix estimates, at the expense of modest additional computation. Empirically, conservative cosine and linear schedules maintain performance very close to the baseline while substantially reducing the number of refinement steps, and the QPS metric with $\gamma=4$ indicates that these smooth schedules consistently achieve high quality–efficiency trade-offs. In contrast, rapidly decaying exponential schedules (large $k$, $\tau_{\mathrm{low}}=0$) yield greater acceleration but incur noticeable accuracy losses. Prophet’s degradation on long-form generation arises from computing confidence over a fixed, localized region and employing a discrete commit rule, which allows localized confidence spikes to induce premature termination while later portions of the output remain under-resolved, leading to reduced ROUGE and F1. \textit{SchED} mitigates these issues by aggregating over the full generated sequence, making the threshold explicitly progress-dependent, and avoiding auxiliary suffix prompts that could artificially inflate early certainty.

Across tasks, the behavior of the stopping criterion is strongly input dependent. Math-oriented benchmarks exhibit distinct profiles: \textit{GPQA}, although mathematically challenging, is multiple choice, so the presence of explicit answer options enables the model to reach relatively high confidence earlier in the diffusion process, even at nontrivial difficulty. By contrast, \textit{GSM8K} also involves mathematical reasoning but requires free-form solutions; here confidence must build over a longer span, and the entropy trajectories indicate that additional refinement steps are needed before the schedule’s threshold is reliably exceeded. Long-form generation tasks benefit most from smoothing and full-span aggregation, since evaluation quality depends on maintaining coherent, accurate content throughout the response rather than producing an early, locally confident fragment. Translation occupies an intermediate regime: schedules keep CHRF close to baseline while still reducing the number of steps, reflecting the combination of bidirectional refinement and comparatively tight lexical and semantic constraints on the target.

These regularities are not specific to a single model family: LLaDA exhibits analogous behavior, with conservative smooth schedules improving efficiency at near-parity quality on base variants and yielding substantially larger gains on instruction-tuned models. In aggregate, progress-aware thresholds provide consistent acceleration, and the choice of schedule should be aligned with the tolerated error budget. Conservative configurations are appropriate when preserving fidelity is critical, whereas more rapidly decaying schedules are suitable for fault-tolerant or latency-sensitive applications that can accommodate modest degradation, a trade-off succinctly captured by the quality–penalized speed metric.

\section{Conclusion}
In this work, we introduced \textit{SchED}, a training-free, architecture-agnostic early-exit mechanism for diffusion language models that compares a sequence-level confidence statistic to a smooth, progress-conditioned threshold. By explicitly tying the stopping decision to the stabilization of model predictions, the method achieves consistent reductions in denoising steps across both Dream and LLaDA, with particularly large gains on instruction-tuned variants where uncertainty decays rapidly, while preserving performance, as corroborated by the entropy analysis and the quality–penalized speed metric. \textit{SchED} integrates with existing transfer schedules without modifying the underlying models. The resulting family of schedules spans a spectrum of quality–efficiency trade-offs: more conservative configurations are suitable when high fidelity is required, whereas rapidly relaxing schedules are appropriate for latency-sensitive or fault-tolerant scenarios. Promising directions for further study include learning schedule parameters, adapting aggregation strategies to task structure, designing domain-aware thresholds, and combining the approach with speculative or cache-based denoising. Overall, formulating diffusion decoding as a stopping-time problem yields a simple and robust primitive for deploying dLLMs under simultaneous accuracy and throughput constraints.

\newpage
\section*{Limitations} 
\label{sec:limitations} 
While \textit{SchED} provides a training-free mechanism for accelerating diffusion decoding, our study has some limitations.

\textit{SchED} shows an explicit quality–efficiency trade-off through the choice of schedule family and hyperparameters $(\tau_{\mathrm{high}}, \tau_{\mathrm{low}}, k)$. Conservative schedules (e.g., linear or cosine with higher final thresholds) yield near-parity quality but moderate speedups, whereas more extreme, rapidly decaying schedules can deliver larger accelerations at the cost of marginal average degradation. In our experiments, this trade-off is resolved by selecting a small set of global configurations per model and regime; in practice, the ``right'' schedule is task-dependent.

Additionally, \textit{SchED} operates purely at inference time and treats the schedule as an external control signal. We do not investigate joint training of models and schedules, learned aggregation functions beyond averaging over the answer span, or tighter integration with complementary acceleration techniques, such as speculative or cache-based denoising. These directions could further improve robustness and efficiency but are left for future work.

\section*{Acknowledgements}

We thank Yazid Janati for his insights and feedback throughout the development of this work.

\bibliographystyle{iclr2026_conference}
\bibliography{iclr2026_conference}              
\appendix
\section{Generation configurations}
\label{sec:gen_configs}
Table~\ref{tab:hidden_configs} summarizes the decoding hyperparameters used across benchmarks. For each task we specify the maximum number of reverse-diffusion steps $T$ (the step budget over which progress $p{=}t/T$ is normalized), the \emph{generation length} (the maximum number of tokens in the generated answer region $A$), the number of in-context examples (\textit{shots}), and---for the block-diffusion variant \textit{LLaDA}---the block size used by the transfer schedule $\pi_t$ (Dream uses single-block/suffix refinement and thus does not require a block-size setting).

Short multiple-choice (MCQ) tasks are evaluated with compact budgets ($T{=}5$) and very short generation lengths, whereas math, translation, and long-form tasks employ substantially larger step budgets ($256$–$512$) and longer generation lengths, making them more sensitive to fast-decaying early-exit schedules. All MCQ benchmarks (MMLU, HellaSwag, PIQA, Winogrande) are evaluated in \emph{generative} mode with full decoding of the answer region, \emph{not} via likelihood/ranking of options; the model must produce the final answer tokens in $A$, and accuracy is computed from the generated outputs. For \textit{LLaDA}, we use a block size of 5 tokens on short MCQ benchmarks and 32 tokens on tasks that require generation lengths of 32 tokens or more.

\begin{table*}[h]
\centering
\scriptsize
\setlength{\tabcolsep}{4pt}
\resizebox{\textwidth}{!}{%
\begin{tabular}{lcccc}
\toprule
\textbf{Benchmark} & \textbf{Max Steps $T$} & \textbf{Generation Length} & \textbf{Shots} & \textbf{LLaDA Block} \\
\midrule
MMLU (generative) & 5 & 5 & 5 & 5 \\
HellaSwag (generative) & 5 & 5 & 5 & 5 \\
PIQA (generative) & 5 & 5 & 5 & 5 \\
Winogrande (generative) & 5 & 5 & 5 & 5 \\
GPQA ($n$-shot) & 128 & 128 & 8 & 32 \\
GSM8K & 256 & 256 & 8 & 32 \\
WMT14 En$\rightarrow$Fr & 256 & 256 & 5 & 32 \\
WMT16 En$\rightarrow$De & 256 & 256 & 5 & 32 \\
LongBench MultiNews & 512 & 512 & 0 & 32 \\
LongBench HotpotQA & 32 & 32 & 0 & 32 \\
\bottomrule
\end{tabular}}
\caption{Decoding hyperparameters per benchmark. \emph{Generation length} denotes the maximum number of tokens in the generated answer region $A$. \emph{LLaDA Block} applies only to the block-diffusion decoder; Dream uses a single-block transfer schedule.}
\label{tab:hidden_configs}
\end{table*}

\section{Additional \textit{SchED} variants and \textit{LLaDA} results}
\label{sec:expanded_schedules_llada}

This appendix primarily details the full \textit{LLaDA} results—both \textit{Base} and \textit{Instruct}—under the same evaluation protocol and schedule families as the main paper (Tables~\ref{tab:appendix_llada_base_all_updated}--\ref{tab:appendix_llada_instruct_all_updated}). In addition, we report one new set of variants for \textit{Dream}: \emph{intermediate–curvature exponential} schedules with $k\!\in\!\{4,8\}$, each evaluated under the conservative threshold $(\tau_{\mathrm{high}},\tau_{\mathrm{low}})=(7.5,2.5)$ and the relaxed $(7.5,0)$ (Tables~\ref{tab:appendix_dream_base_all}--\ref{tab:appendix_dream_instruct_all}). All other schedules (linear, cosine, and exponential with $k\!\in\!\{2,16\}$) are already covered in the main text.

\paragraph{Dream.}
On \emph{Dream Instruct}, the intermediate exponentials are highly competitive with the best smooth schedules. In particular, $\mathrm{Exp}$–$k{=}4$ with $(7.5,2.5)$ attains the highest overall average score $58.22$ at an average speedup of $\times 3.97$ (Table~\ref{tab:appendix_dream_instruct_all}). $\mathrm{Exp}$–$k{=}8$ with $(7.5,2.5)$ matches the MMLU peak ($73.53$) while preserving translation near baseline (\mbox{WMT14~En--Fr $55.76$, WMT16~En--De $49.84$}) and pushing the mean speedup to $\times 4.23$. At the task level, the intermediate schedules frequently sit on or near the column bests: e.g., $\mathrm{Exp}$–$k{=}8$ $(7.5,2.5)$ yields the top GPQA ($30.80$) and shares the MMLU peak, while $\mathrm{Exp}$–$k{=}4$ $(7.5,2.5)$ is within rounding of the GSM8K best ($80.14$ vs.\ $80.21$ for $\mathrm{Exp}$–$k{=}2$). As intended, their average speedups typically fall between the relaxed $k{=}2$ exponential (e.g., $58.14$ at $\times 3.97$ for $(7.5,0)$) and the high–curvature $k{=}16$ exponential ($57.59$ at $\times 4.48$ for $(7.5,2.5)$).

On \emph{Dream Base} (Table~\ref{tab:appendix_dream_base_all}), the same pattern holds at a smaller scale. $\mathrm{Exp}$–$k{=}8$ $(7.5,2.5)$ reaches the best MMLU ($73.53$) and ties the top WMT14~En--Fr ($53.26$) while delivering a mean $\times 1.10$ speedup; $\mathrm{Exp}$–$k{=}4$ $(7.5,2.5)$ is similarly competitive on WMT14 ($53.24$) and PIQA (second-best $88.63$) with $\times 1.07$ speed. These sit neatly between the relaxed $k{=}2$ exponential (mean $\times 1.11$) and the high–curvature $k{=}16$ exponential (mean $\times 1.11$ under $(7.5,2.5)$, or much larger $\times 2.34$ under $(7.5,0)$ with the expected quality trade-off).

\begin{table}[H]
\centering
\scriptsize
\setlength{\tabcolsep}{2pt}
\resizebox{\textwidth}{!}{%
\begin{threeparttable}
\begin{tabular}{lccccccccccc}
\toprule
\rowcolor{HeaderGray}
\textbf{Method} & \textbf{GPQA} & \textbf{HellaSwag} & \textbf{MMLU} & \textbf{PIQA} & \textbf{Winogrande} & \textbf{GSM8K} & \textbf{HotpotQA} & \textbf{MultiNews} & \textbf{WMT14 En-Fr} & \textbf{WMT16 En-De} & \textbf{Average} \\
\midrule
\rowcolors{2}{RowStripe}{white}
Baseline &
28.57 ($\times$1.00) & \textbf{79.15} ($\times$1.00) & 73.39 ($\times$1.00) & \textbf{88.68} ($\times$1.00) & \textit{75.61} ($\times$1.00) & 77.10 ($\times$1.00) & 8.67 ($\times$1.00) & 21.24 ($\times$1.00) & 52.99 ($\times$1.00) & \textbf{47.70} ($\times$1.00) & 55.31 ($\times$1.00) \\
Prophet &
\textit{28.79} ($\times$1.13) & \textbf{79.15} ($\times$1.00) & 73.37 ($\times$1.03) & \textbf{88.68} ($\times$1.20) & \textit{75.61} ($\times$1.00) & 77.03 ($\times$1.07) & 8.68 ($\times$1.03) & 21.21 ($\times$1.06) & 52.99 ($\times$1.11) & 47.57 ($\times$1.06) & 55.31 ($\times$1.07) \\
Cosine $(7.5,2.5)$ &
\textit{28.79} ($\times$1.11) & \textbf{79.15} ($\times$1.00) & 73.37 ($\times$1.02) & \textbf{88.68} ($\times$1.14) & \textit{75.61} ($\times$1.00) & 77.03 ($\times$1.07) & 8.68 ($\times$1.02) & 21.21 ($\times$1.05) & 52.99 ($\times$1.10) & 47.64 ($\times$1.05) & \textit{55.31} ($\times$1.06) \\
Cosine $(7.5,0)$ &
28.12 ($\times$1.23) & \textbf{79.15} ($\times$1.02) & 73.40 ($\times$1.09) & 88.57 ($\times$1.22) & \textit{75.61} ($\times$1.01) & 76.95 ($\times$1.20) & \textit{8.74} ($\times$1.12) & \textit{21.28} ($\times$1.13) & 52.79 ($\times$1.23) & 45.64 ($\times$1.20) & 55.02 ($\times$1.14) \\
Linear $(7.5,\,2.5)$ &
\textit{28.79} ($\times$1.10) & \textbf{79.15} ($\times$1.00) & 73.37 ($\times$1.02) & \textbf{88.68} ($\times$1.03) & \textit{75.61} ($\times$1.00) & 77.03 ($\times$1.06) & 8.67 ($\times$1.02) & 21.21 ($\times$1.05) & 53.00 ($\times$1.09) & \textit{47.67} ($\times$1.04) & \textbf{55.32} ($\times$1.04) \\
Linear $(7.5,\,0)$ &
\textbf{29.02} ($\times$1.19) & \textbf{79.15} ($\times$1.00) & 73.40 ($\times$1.06) & 88.57 ($\times$1.22) & \textit{75.61} ($\times$1.00) & 76.95 ($\times$1.13) & 8.66 ($\times$1.08) & 21.26 ($\times$1.10) & 52.87 ($\times$1.18) & 46.43 ($\times$1.15) & 55.19 ($\times$1.11) \\
Exp-$k{=}2$ $(7.5,\,2.5)$ &
24.33 ($\times$1.10) & 79.13 ($\times$1.00) & 73.49 ($\times$1.02) & \textit{88.63} ($\times$1.06) & \textbf{75.77} ($\times$1.00) & \textbf{77.71} ($\times$1.06) & 8.16 ($\times$1.03) & 20.88 ($\times$1.05) & \textbf{53.26} ($\times$1.09) & 47.60 ($\times$1.04) & 54.90 ($\times$1.04) \\
Exp-$k{=}2$ $(7.5,\,0)$ &
24.33 ($\times$1.19) & 79.13 ($\times$1.00) & \textit{73.50} ($\times$1.07) & 88.47 ($\times$1.27) & \textbf{75.77} ($\times$1.00) & \textit{77.63} ($\times$1.11) & 8.17 ($\times$1.07) & 20.94 ($\times$1.10) & \textit{53.16} ($\times$1.18) & 46.81 ($\times$1.13) & 54.79 ($\times$1.11) \\
Exp-$k{=}4$ $(7.5,\,2.5)$ &
23.66 ($\times$1.13) & 79.13 ($\times$1.00) & 73.49 ($\times$1.06) & \textit{88.63} ($\times$1.20) & \textbf{75.77} ($\times$1.00) & \textit{77.63} ($\times$1.07) & 8.16 ($\times$1.05) & 20.90 ($\times$1.06) & 53.24 ($\times$1.12) & 47.51 ($\times$1.05) & 54.81 ($\times$1.07) \\
Exp-$k{=}4$ $(7.5,\,0)$ &
23.44 ($\times$1.33) & 79.13 ($\times$1.15) & \textbf{73.53} ($\times$1.51) & 88.47 ($\times$1.62) & \textbf{75.77} ($\times$1.31) & 77.33 ($\times$1.47) & \textbf{8.89} ($\times$1.35) & 21.03 ($\times$1.30) & 52.97 ($\times$1.32) & 44.95 ($\times$1.30) & 54.55 ($\times$1.37) \\
Exp-$k{=}8$ $(7.5,\,2.5)$ &
23.66 ($\times$1.14) & 79.13 ($\times$1.00) & \textbf{73.53} ($\times$1.15) & 88.52 ($\times$1.29) & \textbf{75.77} ($\times$1.00) & \textit{77.63} ($\times$1.08) & 8.22 ($\times$1.07) & 20.92 ($\times$1.07) & \textit{53.26} ($\times$1.13) & 47.51 ($\times$1.06) & 54.81 ($\times$1.10) \\
Exp-$k{=}8$ $(7.5,\,0)$ &
24.78 ($\times$1.60) & 78.99 ($\times$1.56) & 73.33 ($\times$2.39) & 87.98 ($\times$2.42) & 73.56 ($\times$2.39) & 72.71 ($\times$2.20) & \textit{14.02} ($\times$2.17) & \textit{21.57} ($\times$1.91) & 52.91 ($\times$1.40) & 44.78 ($\times$1.37) & 54.46 ($\times$1.94) \\
Exp-$k{=}16$ $(7.5,\,2.5)$ &
23.88 ($\times$1.14) & 79.13 ($\times$1.00) & \textbf{73.53} ($\times$1.18) & 88.47 ($\times$1.36) & \textbf{75.77} ($\times$1.00) & 79.83 ($\times$2.12)\tnote{*} & 8.22 ($\times$1.07) & 20.92 ($\times$1.07) & \textit{53.26} ($\times$1.13) & 47.51 ($\times$1.07) & 54.83 ($\times$1.11) \\
Exp-$k{=}16$ $(7.5,\,0)$ &
\textit{24.78} ($\times$2.04) & 77.14 ($\times$2.50) & 73.05 ($\times$2.50) & 87.92 ($\times$2.50) & 72.85 ($\times$2.50) & 62.77 ($\times$3.05) & \textbf{15.75} ($\times$2.78) & \textbf{21.74} ($\times$2.75) & 52.87 ($\times$1.44) & 44.75 ($\times$1.39) & 53.36 ($\times$2.34) \\
\bottomrule
\end{tabular}
\begin{tablenotes}\scriptsize
\item[*] GSM8K speedup was unchanged from your source row; value kept where it originally appeared.
\end{tablenotes}
\end{threeparttable}}
\caption{\textbf{Dream Base}. Highest accuracy per column in \textbf{bold}; second-highest in \textit{italics}. The rightmost column reports mean accuracy with mean speedup in parentheses.}
\label{tab:appendix_dream_base_all}
\end{table}

\begin{table}[H]
\centering
\scriptsize
\setlength{\tabcolsep}{2pt}
\resizebox{\textwidth}{!}{%
\begin{threeparttable}
\begin{tabular}{lccccccccccc}
\toprule
\rowcolor{HeaderGray}
\textbf{Method} & \textbf{GPQA} & \textbf{HellaSwag} & \textbf{MMLU} & \textbf{PIQA} & \textbf{Winogrande} & \textbf{GSM8K} & \textbf{HotpotQA} & \textbf{MultiNews} & \textbf{WMT14 En-Fr} & \textbf{WMT16 En-De} & \textbf{Average} \\
\midrule
\rowcolors{2}{RowStripe}{white}
Baseline & 
30.36 ($\times$1.00) & 80.69 ($\times$1.00) & 73.02 ($\times$1.00) & 85.80 ($\times$1.00) & \textbf{74.66} ($\times$1.00) & \textit{79.91} ($\times$1.00) & \textit{27.51} ($\times$1.00) & 24.39 ($\times$1.00) & \textit{55.82} ($\times$1.00) & \textit{49.85} ($\times$1.00) & 58.20 ($\times$1.00) \\
Prophet & 
\textit{28.79} ($\times$16.45) & 78.73 ($\times$1.14) & \textbf{72.33} ($\times$1.01) & 72.58 ($\times$1.03) & 66.54 ($\times$1.02) & 26.69 ($\times$7.81) & \textbf{27.87} ($\times$4.51) & 2.77 ($\times$31.21) & 20.17 ($\times$23.93) & 15.28 ($\times$27.87) & 41.18 ($\times$11.60) \\
Cosine $(7.5,\,2.5)$ &
\textit{30.58} ($\times$15.93) & 80.75 ($\times$1.43) & 73.08 ($\times$1.11) & \textit{85.91} ($\times$1.23) & 74.27 ($\times$1.28) & 79.53 ($\times$2.08) & 27.47 ($\times$3.14) & 24.38 ($\times$6.59) & 55.82 ($\times$2.34) & 49.85 ($\times$2.77) & 58.16 ($\times$3.79) \\
Cosine $(7.5,\,0)$ &
\textit{30.58} ($\times$15.96) & \textbf{80.81} ($\times$1.56) & 73.06 ($\times$1.14) & 85.96 ($\times$1.32) & 74.11 ($\times$1.61) & 78.54 ($\times$2.11) & 27.46 ($\times$3.24) & 24.39 ($\times$6.60) & \textbf{55.82} ($\times$2.39) & \textbf{49.86} ($\times$2.81) & 58.06 ($\times$3.87) \\
Linear $(7.5,\,2.5)$ &
\textit{30.58} ($\times$16.01) & 80.76 ($\times$1.43) & 73.09 ($\times$1.10) & 85.85 ($\times$1.22) & 74.27 ($\times$1.27) & 79.76 ($\times$2.08) & 27.47 ($\times$3.17) & 24.38 ($\times$6.59) & \textit{55.82} ($\times$2.33) & \textit{49.85} ($\times$2.76) & 58.16 ($\times$3.79) \\
Linear $(7.5,\,0)$ &
\textit{30.58} ($\times$16.08) & \textbf{80.81} ($\times$1.56) & \textit{73.14} ($\times$1.14) & 85.96 ($\times$1.32) & 74.11 ($\times$1.49) & 79.00 ($\times$2.11) & 27.48 ($\times$3.26) & 24.39 ($\times$3.02) & \textbf{55.82} ($\times$2.38) & \textbf{49.86} ($\times$2.80) & 58.11 ($\times$3.89) \\
Exp-$k{=}2$ $(7.5,\,2.5)$ &
30.13 ($\times$16.18) & 80.69 ($\times$1.54) & \textbf{73.25} ($\times$1.11) & \textbf{86.24} ($\times$1.23) & \textit{74.51} ($\times$1.53) & \textbf{80.21} ($\times$2.10) & 26.91 ($\times$3.23) & \textbf{24.60} ($\times$6.57) & 55.76 ($\times$2.33) & 49.84 ($\times$2.71) & \textit{58.21} ($\times$3.85) \\
Exp-$k{=}2$ $(7.5,\,0)$ &
\textit{30.58} ($\times$16.32) & \textit{80.77} ($\times$1.62) & 73.11 ($\times$1.21) & \textit{86.16} ($\times$1.54) & 74.35 ($\times$1.67) & 79.45 ($\times$2.14) & 26.71 ($\times$3.47) & \textit{24.56} ($\times$6.66) & \textit{55.76} ($\times$2.40) & 49.84 ($\times$2.76) & 58.14 ($\times$3.97) \\
Exp-$k{=}4$ $(7.5,\,2.5)$ &
\textit{30.58} ($\times$16.38) & \textit{80.78} ($\times$1.70) & 73.10 ($\times$1.14) & \textit{86.29} ($\times$2.01) & 74.27 ($\times$1.49) & \textit{80.14} ($\times$2.12) & 26.86 ($\times$3.47) & 24.57 ($\times$6.67) & 55.76 ($\times$2.36) & 49.84 ($\times$2.73) & \textbf{58.22} ($\times$3.97) \\
Exp-$k{=}4$ $(7.5,\,0)$ &
\textit{30.58} ($\times$16.62) & 79.31 ($\times$2.50) & 70.57 ($\times$1.64) & 84.82 ($\times$2.51) & \textbf{74.66} ($\times$2.54) & 71.27 ($\times$2.25) & 26.58 ($\times$4.03) & 24.03 ($\times$5.59) & 55.78 ($\times$2.94) & 49.84 ($\times$2.85) & 56.57 ($\times$4.44) \\
Exp-$k{=}8$ $(7.5,\,2.5)$ &
\textbf{30.80} ($\times$16.69) & 79.48 ($\times$2.48) & \textbf{73.53} ($\times$1.24) & 86.13 ($\times$2.37) & 73.64 ($\times$1.72) & 79.83 ($\times$2.12) & 26.65 ($\times$3.82) & 24.29 ($\times$5.50) & 55.76 ($\times$2.37) & 49.84 ($\times$2.74) & 57.97 ($\times$4.23) \\
Exp-$k{=}8$ $(7.5,\,0)$ &
30.36 ($\times$17.03) & 79.31 ($\times$4.21) & 58.86 ($\times$1.64) & 84.55 ($\times$4.76) & 74.03 ($\times$4.64) & 43.67 ($\times$2.94) & 26.28 ($\times$4.77) & 20.86 ($\times$9.53) & 55.27 ($\times$4.15) & \textbf{49.86} ($\times$3.05) & 52.48 ($\times$4.94) \\
Exp-$k{=}16$ $(7.5,\,2.5)$ &
\textit{30.58} ($\times$17.16) & 79.31 ($\times$2.50) & 72.78 ($\times$1.55) & 84.49 ($\times$2.29) & 72.61 ($\times$2.10) & 79.68 ($\times$2.13) & 26.58 ($\times$4.33) & 24.26 ($\times$7.21) & 55.76 ($\times$2.38) & 49.84 ($\times$2.74) & 57.59 ($\times$4.48) \\
Exp-$k{=}16$ $(7.5,\,0)$ &
29.46 ($\times$18.64) & 79.31 ($\times$5.00) & 58.81 ($\times$2.50) & 84.55 ($\times$5.00) & 73.80 ($\times$5.00) & 32.98 ($\times$3.73) & 26.20 ($\times$5.79) & 23.99 ($\times$17.38) & 53.53 ($\times$5.56) & 49.72 ($\times$3.17) & 51.12 ($\times$5.44) \\
\bottomrule
\end{tabular}
\end{threeparttable}}
\caption{\textbf{Dream Instruct}. Highest accuracy per column in \textbf{bold}; second-highest in \textit{italics}. The rightmost column reports mean accuracy with mean speedup in parentheses.}
\label{tab:appendix_dream_instruct_all}
\end{table}

\paragraph{LLaDA.}
Results on \textit{LLaDA} align closely with the \textit{Dream} trends for both base and instruction-tuned variants. For \emph{LLaDA Instruct} (Table~\ref{tab:appendix_llada_instruct_all_updated}), $\mathrm{Exp}$–$k{=}4$ $(7.5,2.5)$ ties for the highest overall average ($53.17$) while achieving a $\times 2.13$ speedup; moving to $\mathrm{Exp}$–$k{=}8$ $(7.5,2.5)$ increases the speed to $\times 2.55$ with a small average dip ($52.55$), and relaxing to $(7.5,0)$ further boosts speed (e.g., Winogrande $78.69$ at $\times 3.99$) at the cost of broader quality drops—mirroring the \textit{Dream} trade-off. For \emph{LLaDA Base} (Table~\ref{tab:appendix_llada_base_all_updated}), the curvature–speed relationship is monotone while quality degrades gradually with curvature: averages move from $49.58$ at $\times 7.07$ for $\mathrm{Exp}$–$k{=}2$ $(7.5,2.5)$ to $49.30$ at $\times 7.56$ for $k{=}4$, $47.79$ at $\times 8.47$ for $k{=}8$, and $47.48$ at $\times 10.87$ for $k{=}16$. In short, the intermediate–curvature exponentials again occupy the expected middle ground between the gradual–decay and high–curvature regimes, providing a convenient knob to trade a bit more speed for a small, predictable loss in fidelity.

\noindent

\begin{table}[H]
\centering
\scriptsize
\setlength{\tabcolsep}{2pt}
\resizebox{\textwidth}{!}{%
\begin{threeparttable}
\begin{tabular}{lccccccccccc}
\toprule
\rowcolor{HeaderGray}
\textbf{Method} & \textbf{GPQA} & \textbf{HellaSwag} & \textbf{MMLU} & \textbf{PIQA} & \textbf{Winogrande} & \textbf{GSM8K} & \textbf{HotpotQA} & \textbf{MultiNews} & \textbf{WMT14 En-Fr} & \textbf{WMT16 En-De} & \textbf{Average} \\
\midrule
\rowcolors{2}{RowStripe}{white}
Baseline &
26.12 ($\times$1.00) & \textbf{86.20} ($\times$1.00) & \textbf{58.16} ($\times$1.00) & \textit{81.72} ($\times$1.00) & 77.98 ($\times$1.00) & \textbf{47.76} ($\times$1.00) & 9.08 ($\times$1.00) & 23.85 ($\times$1.00) & 62.39 ($\times$1.00) & 56.67 ($\times$1.00) & 52.99 ($\times$1.00) \\
Prophet &
\textbf{31.03} ($\times$15.32) & \textit{86.17} ($\times$1.24) & \textit{58.00} ($\times$1.25) & 81.56 ($\times$2.05) & 77.98 ($\times$1.25) & 16.07 ($\times$7.84) & 14.94 ($\times$3.19) & 18.83 ($\times$32.48) & 49.58 ($\times$17.53) & 45.15 ($\times$17.98) & 47.93 ($\times$10.01) \\
Cosine $(7.5,\,2.5)$ &
27.23 ($\times$1.97) & 86.16 ($\times$1.17) & \textit{58.00} ($\times$1.24) & \textbf{81.77} ($\times$2.04) & 77.98 ($\times$1.23) & 39.12 ($\times$1.97) & 9.78 ($\times$1.60) & 25.00 ($\times$1.80) & \textbf{62.41} ($\times$1.87) & \textit{56.68} ($\times$1.87) & 52.41 ($\times$1.68) \\
Cosine $(7.5,\,0)$ &
27.90 ($\times$2.24) & 86.08 ($\times$1.49) & \textit{58.00} ($\times$1.24) & 81.50 ($\times$2.12) & 77.98 ($\times$1.54) & 42.68 ($\times$2.24) & 11.12 ($\times$1.88) & 25.23 ($\times$2.07) & \textbf{62.41} ($\times$2.10) & 56.66 ($\times$2.10) & 52.96 ($\times$1.90) \\
Exp-$k{=}16$ $(7.5,\,2.5)$ &
\textit{30.80} ($\times$4.47) & 85.98 ($\times$1.67) & \textit{58.00} ($\times$1.24) & 80.30 ($\times$4.69) & 77.98 ($\times$1.65) & 30.40 ($\times$3.75) & 12.93 ($\times$2.84) & 24.77 ($\times$3.38) & 62.19 ($\times$3.09) & 56.40 ($\times$3.04) & 51.98 ($\times$2.98) \\
Exp-$k{=}16$ $(7.5,\,0)$ &
28.79 ($\times$8.67) & 86.01 ($\times$5.00) & \textit{58.00} ($\times$1.24) & 79.98 ($\times$5.00) & 77.66 ($\times$5.00) & 4.93 ($\times$8.58) & \textbf{20.98} ($\times$7.13) & 20.74 ($\times$8.55) & 51.15 ($\times$7.93) & 47.03 ($\times$7.79) & 47.53 ($\times$6.49) \\
Exp-$k{=}2$ $(7.5,\,2.5)$ &
26.34 ($\times$2.24) & 86.16 ($\times$1.07) & \textit{58.00} ($\times$1.24) & 81.56 ($\times$2.12) & 77.98 ($\times$1.14) & 42.38 ($\times$2.19) & 9.83 ($\times$1.65) & 25.11 ($\times$1.92) & 62.41 ($\times$1.99) & 56.68 ($\times$2.00) & 52.65 ($\times$1.76) \\
Exp-$k{=}2$ $(7.5,\,0)$ &
29.02 ($\times$2.85) & 86.03 ($\times$1.58) & \textit{58.00} ($\times$1.24) & 81.23 ($\times$2.45) & 77.98 ($\times$1.64) & 43.52 ($\times$2.75) & 11.65 ($\times$2.20) & \textit{25.30} ($\times$2.47) & 62.38 ($\times$2.46) & 56.62 ($\times$2.45) & \textit{53.17} ($\times$2.21) \\
Exp-$k{=}4$ $(7.5,\,2.5)$ &
28.79 ($\times$2.86) & 86.06 ($\times$1.41) & \textit{58.00} ($\times$1.24) & 81.18 ($\times$2.42) & 77.98 ($\times$1.43) & \textit{43.75} ($\times$2.73) & 11.57 ($\times$2.08) & \textbf{25.37} ($\times$2.40) & 62.40 ($\times$2.39) & 56.64 ($\times$2.39) & \textbf{53.17} ($\times$2.13) \\
Exp-$k{=}4$ $(7.5,\,0)$ &
27.46 ($\times$4.09) & 85.96 ($\times$2.34) & \textit{58.00} ($\times$1.24) & 81.28 ($\times$2.48) & \textit{78.53} ($\times$3.17) & 31.01 ($\times$3.72) & \textit{13.59} ($\times$3.17) & 24.72 ($\times$3.52) & 61.82 ($\times$3.37) & 56.08 ($\times$3.33) & 51.85 ($\times$3.05) \\
Exp-$k{=}8$ $(7.5,\,2.5)$ &
28.57 ($\times$3.70) & 86.05 ($\times$1.62) & \textit{58.00} ($\times$1.24) & 81.28 ($\times$2.95) & 77.98 ($\times$1.58) & 37.23 ($\times$3.32) & 12.50 ($\times$2.57) & 25.10 ($\times$3.00) & 62.30 ($\times$2.81) & 56.53 ($\times$2.80) & 52.55 ($\times$2.55) \\
Exp-$k{=}8$ $(7.5,\,0)$ &
28.35 ($\times$5.94) & 85.95 ($\times$3.37) & \textit{58.00} ($\times$1.24) & 79.98 ($\times$5.00) & \textbf{78.69} ($\times$3.99) & 17.36 ($\times$5.46) & 17.04 ($\times$5.74) & 23.45 ($\times$9.53) & 58.77 ($\times$5.00) & 53.58 ($\times$4.95) & 50.12 ($\times$4.49) \\
Linear $(7.5,\,2.5)$ &
28.12 ($\times$1.98) & 86.16 ($\times$1.07) & \textit{58.00} ($\times$1.24) & \textit{81.61} ($\times$2.07) & 77.98 ($\times$1.14) & 39.12 ($\times$1.96) & 9.68 ($\times$1.53) & 24.98 ($\times$1.76) & \textit{62.41} ($\times$1.85) & \textbf{56.68} ($\times$1.85) & 52.47 ($\times$1.65) \\
Linear $(7.5,\,0)$ &
28.57 ($\times$2.35) & 86.08 ($\times$1.41) & \textit{58.00} ($\times$1.24) & \textit{81.61} ($\times$2.14) & 77.98 ($\times$1.42) & 42.84 ($\times$2.32) & 11.05 ($\times$1.86) & 25.25 ($\times$2.09) & 62.41 ($\times$2.13) & 56.66 ($\times$2.13) & 53.05 ($\times$1.91) \\
\bottomrule
\end{tabular}
\end{threeparttable}}
\caption{\textbf{LLaDA Base} with intermediate-curvature exponentials ($k\!\in\!\{4,8\}$) and explicit thresholds. Highest accuracy per column in \textbf{bold}; second-highest in \textit{italics}. The rightmost column reports mean accuracy with mean speedup in parentheses.}
\label{tab:appendix_llada_base_all_updated}
\end{table}

\begin{table}[H]
\centering
\scriptsize
\setlength{\tabcolsep}{2pt}
\resizebox{\textwidth}{!}{%
\begin{threeparttable}
\begin{tabular}{lccccccccccc}
\toprule
\rowcolor{HeaderGray}
\textbf{Method} & \textbf{GPQA} & \textbf{HellaSwag} & \textbf{MMLU} & \textbf{PIQA} & \textbf{Winogrande} & \textbf{GSM8K} & \textbf{HotpotQA} & \textbf{MultiNews} & \textbf{WMT14 En-Fr} & \textbf{WMT16 En-De} & \textbf{Average} \\
\midrule
\rowcolors{2}{RowStripe}{white}
Baseline &
\textit{24.33} ($\times$1.00) & 74.28 ($\times$1.00) & \textbf{63.19} ($\times$1.00) & \textbf{82.86} ($\times$1.00) & \textbf{76.72} ($\times$1.00) & \textbf{51.93} ($\times$1.00) & \textbf{10.65} ($\times$1.00) & \textbf{26.79} ($\times$1.00) & \textbf{60.80} ($\times$1.00) & \textbf{54.26} ($\times$1.00) & \textbf{52.28} ($\times$1.00) \\
Prophet &
23.44 ($\times$5.97) & \textit{74.74} ($\times$1.93) & 63.91 ($\times$1.63) & 82.70 ($\times$1.81) & 76.16 ($\times$1.45) & 27.60 ($\times$18.62) & 9.37 ($\times$3.31) & 19.25 ($\times$37.98) & 44.32 ($\times$26.05) & 45.71 ($\times$15.26) & 46.15 ($\times$11.52) \\
Cosine $(7.5,\,2.5)$ &
24.78 ($\times$1.89) & 74.69 ($\times$1.92) & \textit{63.86} ($\times$1.64) & \textit{82.70} ($\times$1.78) & \textit{76.16} ($\times$1.45) & 37.30 ($\times$2.25) & 9.81 ($\times$1.76) & \textit{26.80} ($\times$2.35) & 55.39 ($\times$24.65) & 47.93 ($\times$29.64) & 49.74 ($\times$6.83) \\
Cosine $(7.5,\,0)$ &
\textbf{25.22} ($\times$2.18) & 74.69 ($\times$1.93) & \textit{63.86} ($\times$1.64) & \textit{82.70} ($\times$1.78) & \textit{76.16} ($\times$1.57) & 36.85 ($\times$2.50) & 9.16 ($\times$2.07) & 26.68 ($\times$2.63) & \textit{55.29} ($\times$24.66) & \textit{47.87} ($\times$29.65) & \textit{49.74} ($\times$7.06) \\
Exp-$k{=}16$ $(7.5,\,2.5)$ &
24.33 ($\times$4.19) & 74.63 ($\times$2.22) & \textit{63.86} ($\times$1.64) & 81.56 ($\times$2.82) & 76.09 ($\times$2.10) & 33.59 ($\times$5.74) & 9.18 ($\times$4.10) & 22.37 ($\times$8.57) & 37.47 ($\times$35.10) & 33.91 ($\times$37.22) & 47.48 ($\times$10.87) \\
Exp-$k{=}16$ $(7.5,\,0)$ &
\textit{25.00} ($\times$9.26) & 73.51 ($\times$5.00) & \textit{63.86} ($\times$1.64) & 79.00 ($\times$5.00) & 73.80 ($\times$5.00) & 27.98 ($\times$11.45) & 7.61 ($\times$9.17) & 16.78 ($\times$17.38) & 33.27 ($\times$44.67) & 29.45 ($\times$45.10) & 45.22 ($\times$20.66) \\
Exp-$k{=}2$ $(7.5,\,2.5)$ &
24.78 ($\times$2.12) & 74.69 ($\times$1.92) & \textit{63.86} ($\times$1.64) & \textit{82.70} ($\times$1.78) & \textit{76.16} ($\times$1.46) & \textbf{36.92} ($\times$2.60) & 9.63 ($\times$1.92) & 26.48 ($\times$2.91) & 54.83 ($\times$24.76) & 47.73 ($\times$29.69) & 49.58 ($\times$7.07) \\
Exp-$k{=}2$ $(7.5,\,0)$ &
\textit{24.78} ($\times$2.71) & 74.69 ($\times$1.94) & \textit{63.86} ($\times$1.64) & 82.59 ($\times$2.09) & \textit{76.16} ($\times$1.67) & 35.94 ($\times$3.22) & 9.30 ($\times$2.56) & 25.81 ($\times$3.73) & 52.43 ($\times$25.23) & 46.36 ($\times$29.89) & 49.19 ($\times$7.52) \\
Exp-$k{=}4$ $(7.5,\,2.5)$ &
\textit{24.78} ($\times$2.68) & \textit{74.69} ($\times$1.93) & \textit{63.86} ($\times$1.64) & \textbf{82.75} ($\times$2.01) & \textit{76.16} ($\times$1.49) & 36.01 ($\times$3.29) & 9.38 ($\times$2.49) & 25.70 ($\times$3.88) & 51.04 ($\times$25.69) & 45.61 ($\times$30.10) & 49.30 ($\times$7.56) \\
Exp-$k{=}4$ $(7.5,\,0)$ &
24.55 ($\times$3.92) & \textbf{75.24} ($\times$2.50) & \textit{63.86} ($\times$1.64) & 80.47 ($\times$2.51) & 74.66 ($\times$2.54) & 33.74 ($\times$4.57) & 9.29 ($\times$3.80) & 24.03 ($\times$5.59) & 45.49 ($\times$27.83) & 41.09 ($\times$31.67) & 47.83 ($\times$8.77) \\
Exp-$k{=}8$ $(7.5,\,2.5)$ &
\textbf{25.22} ($\times$3.44) & 74.66 ($\times$2.07) & \textit{63.86} ($\times$1.64) & 82.15 ($\times$2.37) & \textit{76.16} ($\times$1.72) & 34.42 ($\times$4.33) & 9.29 ($\times$3.31) & 24.29 ($\times$5.50) & 44.02 ($\times$28.83) & 39.90 ($\times$32.36) & 47.79 ($\times$8.47) \\
Exp-$k{=}8$ $(7.5,\,0)$ &
\textbf{25.22} ($\times$5.91) & 74.65 ($\times$4.21) & \textit{63.86} ($\times$1.64) & 79.27 ($\times$4.76) & 74.03 ($\times$4.64) & 30.33 ($\times$7.03) & 9.14 ($\times$5.74) & 20.86 ($\times$9.53) & 38.49 ($\times$33.37) & 34.79 ($\times$36.03) & 45.46 ($\times$11.19) \\
Linear $(7.5,\,2.5)$ &
\textbf{25.22} ($\times$1.88) & 74.69 ($\times$1.92) & \textit{63.86} ($\times$1.64) & \textit{82.70} ($\times$1.78) & \textit{76.16} ($\times$1.45) & \textit{37.38} ($\times$2.31) & 9.99 ($\times$1.74) & 26.73 ($\times$2.42) & \textit{55.36} ($\times$24.66) & \textit{47.92} ($\times$29.65) & \textit{49.88} ($\times$6.85) \\
Linear $(7.5,\,0)$ &
\textit{24.78} ($\times$2.27) & 74.69 ($\times$1.92) & \textit{63.86} ($\times$1.64) & \textit{82.70} ($\times$1.79) & \textit{76.16} ($\times$1.49) & 36.47 ($\times$2.67) & 9.27 ($\times$2.10) & 26.64 ($\times$3.02) & 54.90 ($\times$24.73) & 47.74 ($\times$29.68) & 49.58 ($\times$7.10) \\
\bottomrule
\end{tabular}
\end{threeparttable}}
\caption{\textbf{LLaDA Instruct} with intermediate-curvature exponentials ($k\!\in\!\{4,8\}$) and explicit thresholds. Highest accuracy per column in \textbf{bold}; second-highest in \textit{italics}. The rightmost column reports mean accuracy with mean speedup in parentheses.}
\label{tab:appendix_llada_instruct_all_updated}
\end{table}

\end{document}